\documentclass[11pt]{article}

\usepackage[preprint]{acl}

\usepackage{times}
\usepackage{latexsym}

\usepackage[T1]{fontenc}

\usepackage[utf8]{inputenc}

\usepackage{microtype}

\usepackage{inconsolata}

\usepackage{graphicx}

\usepackage{amsmath}
\usepackage{booktabs}
\usepackage{subcaption}
\usepackage{hyperref}
\usepackage{cleveref}
\usepackage{float}
\usepackage[section]{placeins}
\usepackage[normalem]{ulem}

\usepackage[most]{tcolorbox}
\usepackage{xcolor}

\newtcolorbox{promptbox}{
  colback=gray!4,
  colframe=gray!55,
  boxrule=0.5pt,
  arc=2mm,
  left=1.5mm,
  right=1.5mm,
  top=1mm,
  bottom=1mm,
  fonttitle=\bfseries,
  title=Prompt,
  breakable=false
}

\title{Emulate or Estimate? The Divergent Strengths of Base and Post-Trained Language Models for Opinion Simulation}

\author{
 \textbf{Seth Grief-Albert\textsuperscript{1*}},
 \textbf{Jessica Bo\textsuperscript{2}},
 \textbf{Difan Jiao\textsuperscript{2}},
 \textbf{Ashton Anderson\textsuperscript{2}}
\\
\\
 \textsuperscript{1}Queen's University,
 \textsuperscript{2}University of Toronto
\\
 \textsuperscript{*}
 \small{
   \textbf{Correspondence:} \href{mailto:sethga@cs.toronto.edu}{sethga@cs.toronto.edu} (work done while at the University of Toronto)
 }
}

\begin{document}
\maketitle
\begin{abstract}
Large language models are increasingly used to simulate human opinions, but prior work reports conflicting results: some studies find promising alignment with human survey data, while others find persona collapse and weak demographic sensitivity. We propose that much of this conflict stems from conflating two distinct tasks. We call the first task emulation, in which models generate individual responses that aggregate into a population distribution. We call the second task estimation, in which models directly predict the population distribution. Evaluating six matched base and post-trained models on the Pew American Trends Panel, we find that base models are the stronger emulators: they produce response distributions closer to human ground truth and better preserve demographic structure. Post-trained models are generally the stronger estimators, producing more accurate distributional predictions when asked directly. We argue that model selection for human simulation should be guided by whether the task requires generating text or predicting distributions.
\end{abstract}

\section{Introduction}

\begin{figure}[h!]
    \centering
    \includegraphics[width=\columnwidth]{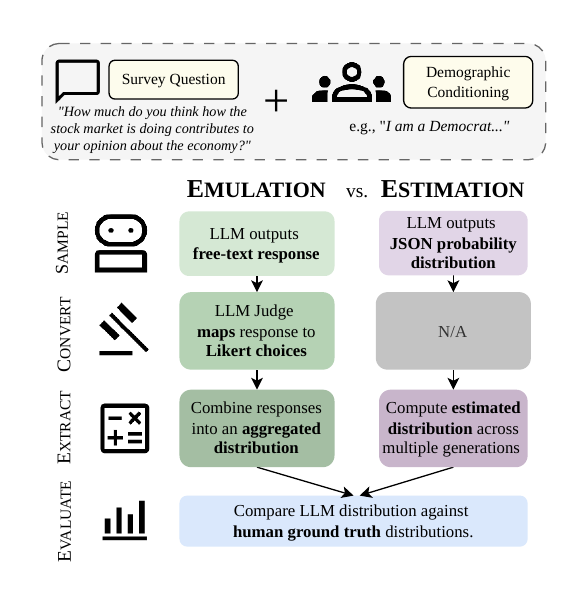}
    \caption{Overview of the evaluation method for \textbf{emulation} and \textbf{estimation}.}
    \label{fig:methods}
\end{figure}

Large Language Models (LLMs), trained on large corpora of human text, are well-positioned to act as simulators of human opinion \citep{argyle-out-of-one-many, durmus2024towards}. High-fidelity population modelling would enable applications across policy testing, market research, and social science, motivating a growing body of work on LLM-based opinion simulation \citep{cao-etal-2025-specializing}. However, prior results are conflicted: some works report promising distributional alignment with human survey data using base models \citep{suh-etal-2025-language, moon-etal-2024-virtual}, while others document persona collapse \citep{li2025llm}, demographic insensitivity \citep{sun-etal-2025-sociodemographic}, and sensitivity to methodological choices \citep{dominguez-olmedo2024questioning, wang-etal-2024-answer-c}. Studies reporting negative results evaluate primarily on post-trained models.

We suggest that this disagreement reflects an underlying conflation of two distinct simulation tasks: \textit{emulation,} in which the model generates responses as an individual participant and the population distribution emerges from aggregation, and \textit{estimation,} in which the model explicitly predicts the population's response distribution. We evaluate three matched base/post-trained pairs under both paradigms on the Pew American Trends Panel, a human opinion survey on economics.

In our case study, base models consistently outperform their post-trained counterparts at emulation, while post-trained variants generally perform better at direct estimation. These findings suggest a basis for model selection in opinion simulation: base models suit tasks requiring generated text, while post-trained models suit direct distributional prediction.

\section{Related Work}
\label{sec:related}

\textbf{LLMs as human simulators.} Language models have been increasingly explored as substrates for simulating human behaviour, spanning survey response simulation \citep{argyle-out-of-one-many, cao-etal-2025-specializing, park2024llmagentsgroundedselfreports}, interactive social agents \citep{park2023generativeagentsinteractivesimulacra, vezhnevets2023generativeagentbasedmodelingactions}, and synthetic population modelling \citep{sun2024randomsiliconsamplingsimulating}. These efforts share the premise that pretraining corpora encode sufficient human diversity to support behavioural simulation at the population level. Following prior work \citep{santurkar-whose-opinions, suh-etal-2025-language}, we evaluate on survey opinion distributions as a controlled setting with ground truth.

\textbf{Negative results and methodological concerns.} A parallel literature has documented significant limitations. Recent works find that post-trained models exhibit systematic persona collapse, producing homogeneous outputs across prompted identities \citep{li2025llm, xiao2026chameleonslimitinvestigatingpersona, qin2026restoringheterogeneityllmbasedsocial}. \citet{sun-etal-2025-sociodemographic} concluded that sociodemographic prompting does not reliably shift model outputs toward target demographics. Methodologically, \citet{tjuatja-etal-2024-llms}, \citet{dominguez-olmedo2024questioning}, and \citet{wang-etal-2024-answer-c} demonstrated that model responses are sensitive to opinion ordering and label assignment. \citet{hullman2026humanstudydidinvolve} argue that the field remains bottlenecked on prediction quality. These studies evaluate primarily or entirely on post-trained models.

\textbf{Base models and distributional measurement.} Several independent findings motivate the present work. \citet{moon-etal-2024-virtual}, \citet{suh-etal-2025-language}, and \citet{cao-etal-2025-specializing} explicitly chose base models, citing opinion-skew in post-trained variants. \citet{santurkar-whose-opinions} observed that base models show closer distributional alignment with human survey data. Separately, \citet{meister-etal-2025-benchmarking} showed that models can describe opinion distributions more accurately than they can produce them through sampling. We synthesize these findings by introducing the emulation-estimation distinction, systematically comparing matched base and post-trained pairs, and proposing open-response emulation as a positional-bias-free evaluation method.

\section{Methodology}

We study the problem of recovering human opinion distributions from language models: given a survey question and an optional demographic condition, produce a distribution over answer options that matches the empirical response distribution of the target population. Survey opinion data provides a controlled evaluation setting with ground-truth distributions across known demographic subgroups, making it a natural testbed for population simulation capabilities. We distinguish two paradigms for this task, and evaluate each on matched base and post-trained model pairs. See \Cref{fig:methods} for the evaluation flowchart.

\subsection{Emulation Paradigm}

Emulation recovers a population distribution by aggregating individual model responses. We implement this via open-response generation: the model produces free-text interview responses without access to multiple-choice options, and an LLM judge maps each response to an answer category (Cohen's \(\kappa=0.66\) with an expert annotator, with full validation in Appendix~\ref{sec:appendix-emulation-llm-judge}). For base models, generation is structured as an interviewer-participant dialogue, with demographic conditioning introduced via a preceding dialogue turn. Post-trained models are given a system instruction to simulate a participant of a given demographic. We compare emulated distributions against human ground truth and a uniform reference. Prompt scaffolds for emulation can be found in Appendix~\ref{sec:appendix-emulation-setup}.

We choose the open-response sampling method over first-token probability extraction, in which the model's logit distribution over answer tokens serves as a direct distributional representation. While this has been widely used across prior works \citep{santurkar-whose-opinions, cao-etal-2025-specializing, suh-etal-2025-language}, it is confounded by positional bias \citep{wang-etal-2024-answer-c, tjuatja-etal-2024-llms}. Moreover, if the goal of emulation is to produce text that can be validated, first-token extraction offers no interpretable output. We adopt open-response emulation to eliminate the positional confounder, and present the first-token results --- which corroborate our main findings --- as Appendix~\ref{sec:appendix-first-token-extraction}.

\subsection{Estimation Paradigm}

In the estimation paradigm, the model directly predicts the population's distribution for a given survey question, following the verbalized distribution approach introduced by \citet{meister-etal-2025-benchmarking}. The model receives the question with multiple-choice options and is prompted to return a JSON object representing the estimated probability distribution over answers. Prompt scaffolds for estimation can be found in Appendix~\ref{sec:appendix-estimation-setup}.

Post-trained models produce stable outputs under greedy decoding, so a single sample is taken. Base model completions are more variable, so we sample multiple completions and average the resulting distributions. We avoid providing a few-shot example format so as not to influence the prior. Both model types are prompted with appropriate formats to estimate the response distribution of a given demographic group.

\subsection{Metrics}

We evaluate distributional fidelity using two complementary metrics. Total Variation Distance (TVD) measures distributional divergence without regard to ordinality. Wasserstein distance respects the ordinal structure of answer options, penalizing errors proportionally to their distance from the correct response. Both metrics are standard in the distributional alignment literature \citep{santurkar-whose-opinions, moon-etal-2024-virtual, meister-etal-2025-benchmarking}. Full definitions are in Appendix~\ref{sec:appendix-metric-definitions}.

\section{Experiments}

\begin{figure*}[t]
\centering

\begin{subfigure}[t]{0.58\linewidth}
  \centering
  \includegraphics[width=\linewidth]{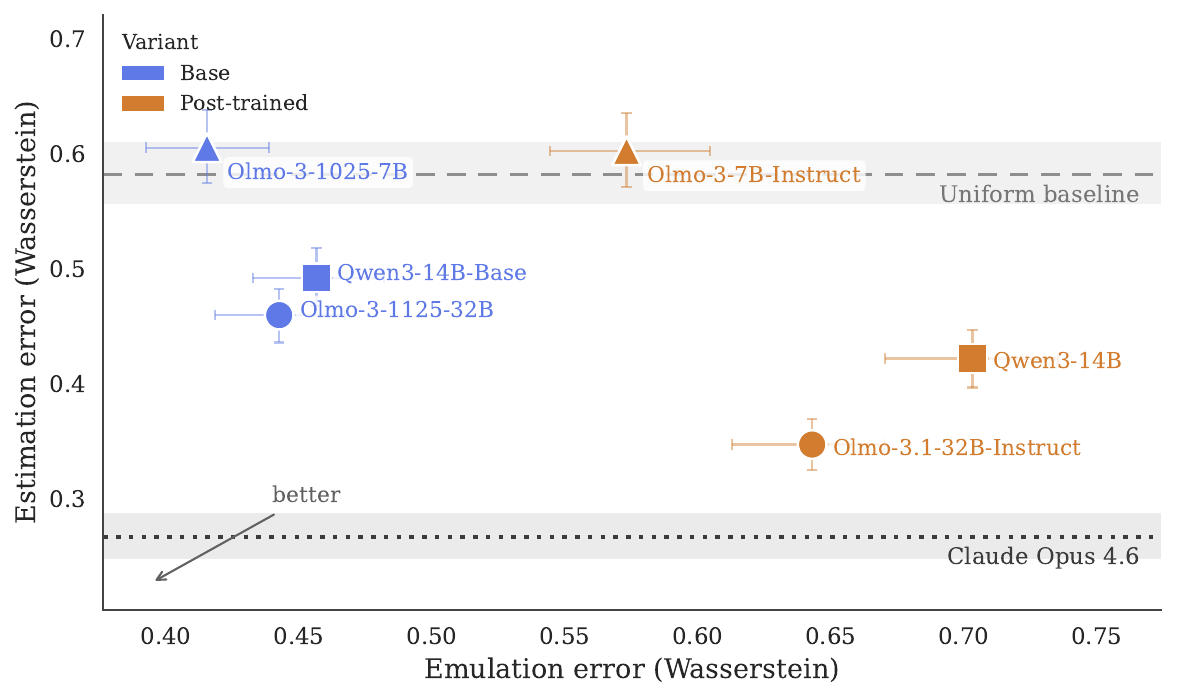}
  \caption{Emulation-estimation tradeoff across base and post-trained models. Each point represents mean Wasserstein distance against human ground-truth distributions, averaged across seven conditions. Lower is better on both axes. Error bars: 95\% bootstrap CIs from question resampling (2,000 replicates). Baselines shown with uncertainty bands.}
  \label{fig:wasserstein-scatter}
\end{subfigure}
\hfill
\begin{subfigure}[t]{0.34\linewidth}
  \centering
  \includegraphics[width=\linewidth]{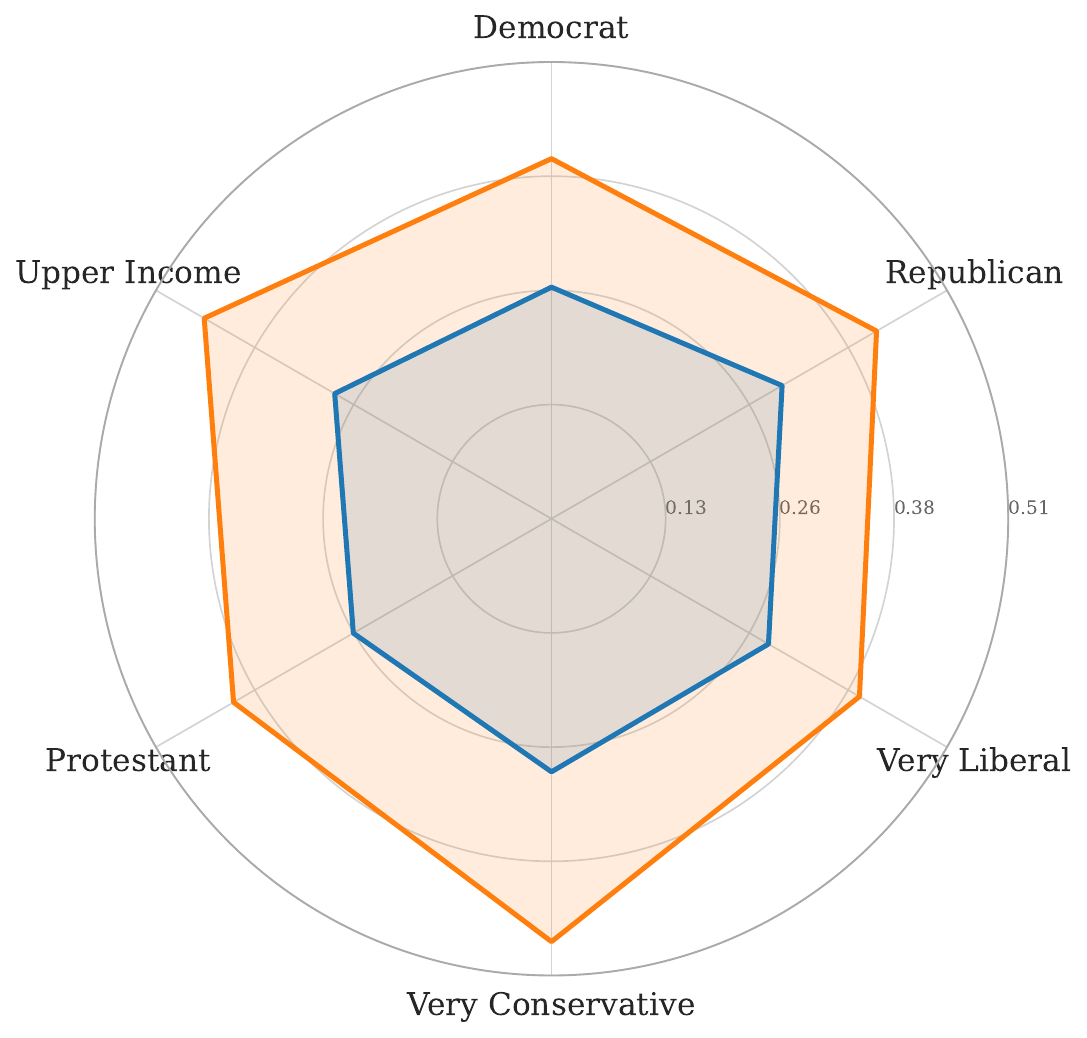}
  \caption{Per-condition emulation error (TVD) for Olmo-3-1125-32B (base, blue) and Olmo-3.1-32B-Instruct (post-trained, orange). Smaller area indicates closer alignment with human distributions.}
  \label{fig:olmo-radar}
\end{subfigure}

\caption{Summary of model error patterns. Base models are blue and post-trained models are orange.}
\label{fig:model-error-graphics}
\end{figure*}

\subsection{Experimental Setup}

Our evaluation dataset consists of 59 four-option items from the Pew Research Center's American Trends Panel Wave 54 \citep{pew-atp-wave54-2019}, which specifically elicits economic opinions. Refusal responses are dropped and human distributions renormalized. We apply seven demographic conditions: an unconditioned marginal and six demographics spanning political (Democrat, Republican), ideological (Very Liberal, Very Conservative), economic (Upper-income), and religious (Protestant) axes. We evaluate six open-weight models with matched base and post-trained variants: Qwen3-14B, Olmo-3-7B, and Olmo-3-32B \citep{yang2025qwen3technicalreport, olmo2026olmo3}. Post-trained models are sampled at temperature 1.5 in emulation to support output diversity. We also include Claude Opus 4.6 \citep{anthropic2026opus46} as a frontier estimation reference.

\subsection{Base Models are Emulators}

\begin{table}[t]
\centering
\scriptsize
\caption{Average error across the seven evaluation conditions. Lower is better.}
\label{tab:model-error}
\begin{tabular}{lrr|rr}
\toprule
& \multicolumn{2}{c|}{Estimation error} & \multicolumn{2}{c}{Emulation error} \\
\cmidrule(lr){2-3}
\cmidrule(lr){4-5}
Model & TVD & Wasserstein & TVD & Wasserstein \\
\midrule
Qwen3-14B Base & 0.236 & 0.492 & \textbf{0.263} & \textbf{0.457} \\
Qwen3-14B & \textbf{0.213} & \textbf{0.422} & 0.458 & 0.703 \\
\midrule
Olmo-3-1025-7B & \textbf{0.278} & 0.605 & \textbf{0.248} & \textbf{0.415} \\
Olmo-3-7B-Instruct & 0.285 & \textbf{0.603} & 0.372 & 0.573 \\
\midrule
Olmo-3-1125-32B & 0.223 & 0.460 & \textbf{0.274} & \textbf{0.443} \\
Olmo-3.1-32B-Instruct & \textbf{0.185} & \textbf{0.347} & 0.427 & 0.643 \\
\midrule
Claude Opus 4.6 & \textbf{0.142} & \textbf{0.267} & — & — \\
\midrule
uniform & 0.271 & 0.582 & — & — \\
\bottomrule
\end{tabular}
\end{table}

Base models produce emulated distributions consistently closer to human ground truth than their post-trained counterparts across all six models and all seven evaluation conditions (\Cref{tab:model-error}). On Wasserstein distance, base models outperform their post-trained counterparts in every setting and beat the uniform reference in 20 of 21 model-condition comparisons, indicating that even when base models miss exact categorical probabilities, mass is placed near the correct ordered responses. On TVD, base models again outperform their post-trained counterparts in every setting, though they beat the uniform reference distribution in only 11 of 21 comparisons --- a weaker margin than Wasserstein, where ordinal proximity to correct answers is rewarded. The per-condition comparison of Olmo-32B base and post-trained variants confirms this advantage is consistent across all six demographic conditions (\Cref{fig:olmo-radar}). This gap is driven by mode collapse in post-trained outputs, which are \(8.0\)--\(14.0\times\) more similar at the bigram level than base model outputs (\Cref{tab:open-response-jaccard}); elevated temperature does not resolve this collapse.

To test whether conditioned models track genuine demographic structure, we correlate model and human pairwise TVD distances among the six conditioned groups (15 pairs) via Spearman's \(\rho\) (\Cref{tab:spearman_correlations}). Base models consistently align with human demographic structure (\(\rho=0.61\)--\(0.75\), all \(p<0.02\)) while compressing inter-group differences to approximately 70\% of human magnitude. Post-trained models show weaker structural alignment (\(\rho=0.35\)--\(0.59\)) and exaggerate inter-group differences by approximately 2\(\times\), consistent with stereotyped persona simulation \citep{li2025llm}.

\begin{table}[h]
\centering
\small
\caption{Spearman correlations and mean ratio of inter-group distribution between open-response model and human pairwise distances.}
\label{tab:spearman_correlations}
\begin{tabular}{lccc}
\toprule
Model & $\rho$ & $p$ & mean ratio\\
\midrule
Qwen3-14B Base        & 0.754 & 0.001 & 0.688\\
Qwen3-14B             & 0.536 & 0.040 & 2.242\\
\midrule
Olmo-3-1025-7B        & 0.682 & 0.005 & 0.664\\
Olmo-3-7B-Instruct    & 0.350 & 0.201 & 2.014\\
\midrule
Olmo-3-1125-32B       & 0.607 & 0.016 & 0.704\\
Olmo-3.1-32B-Instruct & 0.593 & 0.020 & 2.063\\
\bottomrule
\end{tabular}
\end{table}
    
\subsection{Post-trained Models are Estimators}

Under the estimation paradigm, post-trained models produce distributions closer to human ground truth than the uniform baseline in most conditions (\Cref{tab:model-error}). Estimation error decreases with model scale and is generally lower in post-trained models than their base counterparts, with the 7B pair as an exception. Claude Opus 4.6 achieves the strongest performance. This is consistent with \citet{meister-etal-2025-benchmarking}, who find that distributional knowledge is accessible through verbalization.

Base models also produce estimation results that match or exceed the uniform baseline in most configurations, indicating that distributional knowledge about human opinion is present prior to post-training.

\section{Discussion}

The results of our case study suggest that opinion simulation should be evaluated according to the form of output required by the downstream task. If the goal is to estimate aggregate opinion distributions, post-trained models are the more appropriate substrate. If the goal is to generate text that aligns with a population's diversity --- as would be required for synthetic respondents or interactive agents --- base models are more directionally aligned with the task, because the desired distribution must emerge from sampled outputs rather than from a direct verbal estimate. More broadly, evaluations should measure whether generated responses preserve both aggregate opinion distributions and demographic structure, rather than only whether a model can describe those distributions when prompted directly.

This distinction also clarifies why prior findings on LLM opinion simulation appear mixed. Evaluations based on explicit prediction and evaluations based on generated responses measure different capabilities. Post-training appears to improve distributional estimation while reducing the output diversity needed for faithful emulation. The Persona Selection Model \citep{marks2026persona} offers one theoretical account of this tradeoff. Post-training steers the base model's broad, chimeric distribution into a single Assistant persona optimized for helpfulness \citep{lu2026assistantaxissituatingstabilizing}. When asked to emulate a demographic group, the post-trained model may be performing \textit{second-order simulation} --- an Assistant persona simulating a demographic persona --- that compresses output diversity through the bottleneck of a single character. The Assistant persona produces well-calibrated estimates because the underlying LLM has learned that a helpful assistant, when asked a factual question, will draw on its available knowledge to provide a useful answer \citep{shanahan_role_2023}. Base models may bypass this bottleneck: their outputs draw directly from the pre-trained distribution over personas, which could preserve more of the representative diversity needed for emulation at the cost of weaker metacognitive estimation.

The output restriction we observe is not specific to our task; post-trained models also underrepresent the spread of human viewpoints on open subjective questions \citep{poole-dayan2026benchmarking}. Where in post-training this restriction arises remains an open empirical question. Characterizing it mechanistically would mean identifying at which stage emulative output diversity is lost, and whether conditioned generations route through shared internal representations. This would also distinguish the persona-bottleneck account from a simpler explanation in which post-training reduces output entropy without any persona-level structure.

\section{Conclusion}

We distinguish two paradigms for simulating human opinion distributions with language models: emulation and estimation. In our case study, post-trained models excel at estimation, producing well-calibrated distributional predictions when asked explicitly. Base models are stronger at emulation, generating response distributions that more faithfully track human demographic structure: inter-group differences are compressed to approximately 70\% of human magnitude, compared to the approximately 2\(\times\) exaggeration observed in post-trained models. We propose open-response emulation as an evaluation paradigm free of the positional bias that confounds first-token extraction, and argue that the choice of simulation substrate should be guided by whether the downstream task requires distributional estimates or generated text.

\section*{Limitations}

Whether the emulation-estimation tradeoff persists or collapses at frontier scale is unknown, as frontier base model variants are not released. Post-training pipelines also vary substantially across model families in ways that may modulate this tradeoff \citep{10.1145/3805689.3806444}.

A second limitation concerns temporal alignment. The Pew survey data was collected in 2019, while the models were trained on corpora extending through 2024--2025. Human opinion distributions shift over time, and it is unclear whether models simulate the opinions of a particular era or encode a temporally blended representation. This confound applies to both base and post-trained models. Whether effective population simulation will require continually updated models, or whether temporal dynamics can be controlled for, remains an open question requiring community-wide effort.

Finally, our evaluation is limited to survey opinion; extending the capabilities of base models to richer behavioural settings, including multi-turn dialogue and agent-based simulation, is a natural step toward the broader goal of faithful population modelling. Our grounding in economic opinion as a case study domain can also be extended to include other domains, such as immigration, climate, and health policy. Domains where opinion is less strongly partisan-sorted would test whether the demographic structure base models recover generalizes beyond politically coded attitudes.

\section*{Ethical Considerations}

Simulating demographic opinion distributions requires conditioning on broad categories, which risks reinforcing stereotypes if models collapse groups to caricatured responses. We adopt distributional evaluation, measuring how well models capture the full spread of opinion within groups rather than a single modal response. While we argue that base models are a more suitable substrate for text-generating simulation, we caution that any deployment as synthetic respondents should be validated against real population data before informing decisions. All human ground-truth data is drawn from the Pew Research Center's American Trends Panel, a consented and anonymized survey instrument.


\bibliography{custom}

@misc{marks2026persona,
  author = {Marks, Sam and Lindsey, Jack and Olah, Christopher},
  title = {The Persona Selection Model: Why {AI} Assistants Might Behave Like Humans},
  year = {2026},
  month = feb,
  day = {23},
  howpublished = {\url{https://alignment.anthropic.com/2026/psm/}},
}

@inproceedings{santurkar-whose-opinions,
author = {Santurkar, Shibani and Durmus, Esin and Ladhak, Faisal and Lee, Cinoo and Liang, Percy and Hashimoto, Tatsunori},
title = {Whose opinions do language models reflect?},
year = {2023},
publisher = {JMLR.org},
booktitle = {Proceedings of the 40th International Conference on Machine Learning},
articleno = {1244},
numpages = {34},
location = {Honolulu, Hawaii, USA},
series = {ICML'23}
}

@inproceedings{meister-etal-2025-benchmarking,
    title = "Benchmarking Distributional Alignment of Large Language Models",
    author = "Meister, Nicole  and
      Guestrin, Carlos  and
      Hashimoto, Tatsunori",
    editor = "Chiruzzo, Luis  and
      Ritter, Alan  and
      Wang, Lu",
    booktitle = "Proceedings of the 2025 Conference of the Nations of the Americas Chapter of the Association for Computational Linguistics: Human Language Technologies (Volume 1: Long Papers)",
    month = apr,
    year = "2025",
    address = "Albuquerque, New Mexico",
    publisher = "Association for Computational Linguistics",
    url = "https://aclanthology.org/2025.naacl-long.2/",
    doi = "10.18653/v1/2025.naacl-long.2",
    pages = "24--49",
    ISBN = "979-8-89176-189-6"
}

@inproceedings{cao-etal-2025-specializing,
    title = "Specializing Large Language Models to Simulate Survey Response Distributions for Global Populations",
    author = {Cao, Yong  and
      Liu, Haijiang  and
      Arora, Arnav  and
      Augenstein, Isabelle  and
      R{\"o}ttger, Paul  and
      Hershcovich, Daniel},
    editor = "Chiruzzo, Luis  and
      Ritter, Alan  and
      Wang, Lu",
    booktitle = "Proceedings of the 2025 Conference of the Nations of the Americas Chapter of the Association for Computational Linguistics: Human Language Technologies (Volume 1: Long Papers)",
    month = apr,
    year = "2025",
    address = "Albuquerque, New Mexico",
    publisher = "Association for Computational Linguistics",
    url = "https://aclanthology.org/2025.naacl-long.162/",
    doi = "10.18653/v1/2025.naacl-long.162",
    pages = "3141--3154",
    ISBN = "979-8-89176-189-6"
}

@inproceedings{suh-etal-2025-language,
    title = "Language Model Fine-Tuning on Scaled Survey Data for Predicting Distributions of Public Opinions",
    author = "Suh, Joseph  and
      Jahanparast, Erfan  and
      Moon, Suhong  and
      Kang, Minwoo  and
      Chang, Serina",
    editor = "Che, Wanxiang  and
      Nabende, Joyce  and
      Shutova, Ekaterina  and
      Pilehvar, Mohammad Taher",
    booktitle = "Proceedings of the 63rd Annual Meeting of the Association for Computational Linguistics (Volume 1: Long Papers)",
    month = jul,
    year = "2025",
    address = "Vienna, Austria",
    publisher = "Association for Computational Linguistics",
    url = "https://aclanthology.org/2025.acl-long.1028/",
    doi = "10.18653/v1/2025.acl-long.1028",
    pages = "21147--21170",
    ISBN = "979-8-89176-251-0"
}

@article{argyle-out-of-one-many,
author = {Argyle, Lisa and Busby, Ethan and Fulda, Nancy and Gubler, Joshua and Rytting, Christopher and Wingate, David},
year = {2023},
month = {02},
pages = {337–351},
title = {Out of One, Many: Using Language Models to Simulate Human Samples},
volume = {31},
number={3},
journal = {Political Analysis},
doi = {10.1017/pan.2023.2}
}

@inproceedings{park2023generativeagentsinteractivesimulacra,
author = {Park, Joon Sung and O'Brien, Joseph and Cai, Carrie Jun and Morris, Meredith Ringel and Liang, Percy and Bernstein, Michael S.},
title = {Generative Agents: Interactive Simulacra of Human Behavior},
year = {2023},
isbn = {9798400701320},
publisher = {Association for Computing Machinery},
address = {New York, NY, USA},
url = {https://doi.org/10.1145/3586183.3606763},
doi = {10.1145/3586183.3606763},
booktitle = {Proceedings of the 36th Annual ACM Symposium on User Interface Software and Technology},
articleno = {2},
numpages = {22},
location = {San Francisco, CA, USA},
series = {UIST '23}
}

@inproceedings{wang-etal-2024-answer-c,
    title = "``{My} Answer is {C}'': First-Token Probabilities Do Not Match Text Answers in Instruction-Tuned Language Models",
    author = {Wang, Xinpeng  and
      Ma, Bolei  and
      Hu, Chengzhi  and
      Weber-Genzel, Leon  and
      R{\"o}ttger, Paul  and
      Kreuter, Frauke  and
      Hovy, Dirk  and
      Plank, Barbara},
    editor = "Ku, Lun-Wei  and
      Martins, Andre  and
      Srikumar, Vivek",
    booktitle = "Findings of the Association for Computational Linguistics: ACL 2024",
    month = aug,
    year = "2024",
    address = "Bangkok, Thailand",
    publisher = "Association for Computational Linguistics",
    url = "https://aclanthology.org/2024.findings-acl.441/",
    doi = "10.18653/v1/2024.findings-acl.441",
    pages = "7407--7416"
}

@inproceedings{moon-etal-2024-virtual,
    title = "Virtual Personas for Language Models via an Anthology of Backstories",
    author = "Moon, Suhong  and
      Abdulhai, Marwa  and
      Kang, Minwoo  and
      Suh, Joseph  and
      Soedarmadji, Widyadewi  and
      Behar, Eran Kohen  and
      Chan, David M.",
    editor = "Al-Onaizan, Yaser  and
      Bansal, Mohit  and
      Chen, Yun-Nung",
    booktitle = "Proceedings of the 2024 Conference on Empirical Methods in Natural Language Processing",
    month = nov,
    year = "2024",
    address = "Miami, Florida, USA",
    publisher = "Association for Computational Linguistics",
    url = "https://aclanthology.org/2024.emnlp-main.1110/",
    doi = "10.18653/v1/2024.emnlp-main.1110",
    pages = "19864--19897"
}

@inproceedings{sun-etal-2025-sociodemographic,
    title = "Sociodemographic Prompting is Not Yet an Effective Approach for Simulating Subjective Judgments with {LLM}s",
    author = "Sun, Huaman  and
      Pei, Jiaxin  and
      Choi, Minje  and
      Jurgens, David",
    editor = "Chiruzzo, Luis  and
      Ritter, Alan  and
      Wang, Lu",
    booktitle = "Proceedings of the 2025 Conference of the Nations of the Americas Chapter of the Association for Computational Linguistics: Human Language Technologies (Volume 2: Short Papers)",
    month = apr,
    year = "2025",
    address = "Albuquerque, New Mexico",
    publisher = "Association for Computational Linguistics",
    url = "https://aclanthology.org/2025.naacl-short.71/",
    doi = "10.18653/v1/2025.naacl-short.71",
    pages = "845--854",
    ISBN = "979-8-89176-190-2"
}

@inproceedings{
durmus2024towards,
title={Towards Measuring the Representation of Subjective Global Opinions in Language Models},
author={Esin Durmus and Karina Nguyen and Thomas Liao and Nicholas Schiefer and Amanda Askell and Anton Bakhtin and Carol Chen and Zac Hatfield-Dodds and Danny Hernandez and Nicholas Joseph and Liane Lovitt and Sam McCandlish and Orowa Sikder and Alex Tamkin and Janel Thamkul and Jared Kaplan and Jack Clark and Deep Ganguli},
booktitle={First Conference on Language Modeling},
year={2024},
url={https://openreview.net/forum?id=zl16jLb91v}
}

@misc{sun2024randomsiliconsamplingsimulating,
      title={Random Silicon Sampling: Simulating Human Sub-Population Opinion Using a Large Language Model Based on Group-Level Demographic Information}, 
      author={Seungjong Sun and Eungu Lee and Dongyan Nan and Xiangying Zhao and Wonbyung Lee and Bernard J. Jansen and Jang Hyun Kim},
      year={2024},
      eprint={2402.18144},
      archivePrefix={arXiv},
      primaryClass={cs.AI},
      url={https://arxiv.org/abs/2402.18144}, 
}

@misc{park2024llmagentsgroundedselfreports,
      title={{LLM} Agents Grounded in Self-Reports Enable General-Purpose Simulation of Individuals}, 
      author={Joon Sung Park and Carolyn Q. Zou and Jonne Kamphorst and Niles Egan and Aaron Shaw and Benjamin Mako Hill  and Carrie Cai and Meredith Ringel Morris and Percy Liang and Robb Willer and Michael S. Bernstein},
      year={2024},
      eprint={2411.10109},
      archivePrefix={arXiv},
      primaryClass={cs.AI},
      url={https://arxiv.org/abs/2411.10109}, 
}

@inproceedings{
li2025llm,
title={{LLM} Generated Persona is a Promise with a Catch},
author={Ang Li and Haozhe Chen and Hongseok Namkoong and Tianyi Peng},
booktitle={The Thirty-Ninth Annual Conference on Neural Information Processing Systems Position Paper Track},
year={2025},
url={https://openreview.net/forum?id=qh9eGtMG4H}
}

@article{tjuatja-etal-2024-llms,
    author = {Tjuatja, Lindia and Chen, Valerie and Wu, Tongshuang and Talwalkwar, Ameet and Neubig, Graham},
    title = {Do {LLM}s Exhibit Human-like Response Biases? {A} Case Study in Survey Design},
    journal = {Transactions of the Association for Computational Linguistics},
    volume = {12},
    pages = {1011-1026},
    year = {2024},
    month = {09},
    issn = {2307-387X},
    doi = {10.1162/tacl_a_00685},
    url = {https://doi.org/10.1162/tacl_a_00685},
    eprint = {https://direct.mit.edu/tacl/article-pdf/doi/10.1162/tacl_a_00685/2468689/tacl_a_00685.pdf},
}

@inproceedings{
dominguez-olmedo2024questioning,
title={Questioning the Survey Responses of Large Language Models},
author={Ricardo Dominguez-Olmedo and Moritz Hardt and Celestine Mendler-D{\"u}nner},
booktitle={ICLR 2024 Workshop on Reliable and Responsible Foundation Models},
year={2024},
url={https://openreview.net/forum?id=cvy6DMGnqP}
}

@misc{xiao2026chameleonslimitinvestigatingpersona,
      title={The Chameleon's Limit: Investigating Persona Collapse and Homogenization in Large Language Models}, 
      author={Yunze Xiao and Vivienne J. Zhang and Chenghao Yang and Ningshan Ma and Weihao Xuan and {Jen-tse} Huang},
      year={2026},
      eprint={2604.24698},
      archivePrefix={arXiv},
      primaryClass={cs.CL},
      url={https://arxiv.org/abs/2604.24698}, 
}

@misc{hullman2026humanstudydidinvolve,
      title={This human study did not involve human subjects: Validating {LLM} simulations as behavioral evidence}, 
      author={Jessica Hullman and David Broska and Huaman Sun and Aaron Shaw},
      year={2026},
      eprint={2602.15785},
      archivePrefix={arXiv},
      primaryClass={cs.AI},
      url={https://arxiv.org/abs/2602.15785}, 
}

@techreport{anthropic2026opus46,
  author      = {{Anthropic}},
  title       = {Claude {Opus} 4.6 System Card},
  institution = {Anthropic},
  year        = {2026},
  url         = {https://www-cdn.anthropic.com/14e4fb01875d2a69f646fa5e574dea2b1c0ff7b5.pdf}
}

@misc{yang2025qwen3technicalreport,
      title={Qwen3 Technical Report}, 
      author={An Yang and Anfeng Li and Baosong Yang and Beichen Zhang and Binyuan Hui and Bo Zheng and Bowen Yu and Chang Gao and Chengen Huang and Chenxu Lv and Chujie Zheng and Dayiheng Liu and Fan Zhou and Fei Huang and Feng Hu and Hao Ge and Haoran Wei and Huan Lin and Jialong Tang and Jian Yang and Jianhong Tu and Jianwei Zhang and Jianxin Yang and Jiaxi Yang and Jing Zhou and Jingren Zhou and Junyang Lin and Kai Dang and Keqin Bao and Kexin Yang and Le Yu and Lianghao Deng and Mei Li and Mingfeng Xue and Mingze Li and Pei Zhang and Peng Wang and Qin Zhu and Rui Men and Ruize Gao and Shixuan Liu and Shuang Luo and Tianhao Li and Tianyi Tang and Wenbiao Yin and Xingzhang Ren and Xinyu Wang and Xinyu Zhang and Xuancheng Ren and Yang Fan and Yang Su and Yichang Zhang and Yinger Zhang and Yu Wan and Yuqiong Liu and Zekun Wang and Zeyu Cui and Zhenru Zhang and Zhipeng Zhou and Zihan Qiu},
      year={2025},
      eprint={2505.09388},
      archivePrefix={arXiv},
      primaryClass={cs.CL},
      url={https://arxiv.org/abs/2505.09388}, 
}

@misc{olmo2026olmo3,
      title={Olmo 3}, 
      author={Team Olmo and Allyson Ettinger and Amanda Bertsch and Bailey Kuehl and David Graham and David Heineman and Dirk Groeneveld and Faeze Brahman and Finbarr Timbers and Hamish Ivison and Jacob Morrison and Jake Poznanski and Kyle Lo and Luca Soldaini and Matt Jordan and Mayee Chen and Michael Noukhovitch and Nathan Lambert and Pete Walsh and Pradeep Dasigi and Robert Berry and Saumya Malik and Saurabh Shah and Scott Geng and Shane Arora and Shashank Gupta and Taira Anderson and Teng Xiao and Tyler Murray and Tyler Romero and Victoria Graf and Akari Asai and Akshita Bhagia and Alexander Wettig and Alisa Liu and Aman Rangapur and Chloe Anastasiades and Costa Huang and Dustin Schwenk and Harsh Trivedi and Ian Magnusson and Jaron Lochner and Jiacheng Liu and Lester James V. Miranda and Maarten Sap and Malia Morgan and Michael Schmitz and Michal Guerquin and Michael Wilson and Regan Huff and Ronan Le Bras and Rui Xin and Rulin Shao and Sam Skjonsberg and Shannon Zejiang Shen and Shuyue Stella Li and Tucker Wilde and Valentina Pyatkin and Will Merrill and Yapei Chang and Yuling Gu and Zhiyuan Zeng and Ashish Sabharwal and Luke Zettlemoyer and Pang Wei Koh and Ali Farhadi and Noah A. Smith and Hannaneh Hajishirzi},
      year={2026},
      eprint={2512.13961},
      archivePrefix={arXiv},
      primaryClass={cs.CL},
      url={https://arxiv.org/abs/2512.13961}, 
}

@misc{vezhnevets2023generativeagentbasedmodelingactions,
      title={Generative agent-based modeling with actions grounded in physical, social, or digital space using {Concordia}}, 
      author={Alexander Sasha Vezhnevets and John P. Agapiou and Avia Aharon and Ron Ziv and Jayd Matyas and Edgar A. Duéñez-Guzmán and William A. Cunningham and Simon Osindero and Danny Karmon and Joel Z. Leibo},
      year={2023},
      eprint={2312.03664},
      archivePrefix={arXiv},
      primaryClass={cs.AI},
      url={https://arxiv.org/abs/2312.03664}, 
}

@misc{lu2026assistantaxissituatingstabilizing,
      title={The Assistant Axis: Situating and Stabilizing the Default Persona of Language Models}, 
      author={Christina Lu and Jack Gallagher and Jonathan Michala and Kyle Fish and Jack Lindsey},
      year={2026},
      eprint={2601.10387},
      archivePrefix={arXiv},
      primaryClass={cs.CL},
      url={https://arxiv.org/abs/2601.10387}, 
}

@inproceedings{
wallach2025position,
title={Position: Evaluating Generative {AI} Systems Is a Social Science Measurement Challenge},
author={Hanna Wallach and Meera Desai and A. Feder Cooper and Angelina Wang and Chad Atalla and Solon Barocas and Su Lin Blodgett and Alexandra Chouldechova and Emily Corvi and P. Alex Dow and Jean Garcia-Gathright and Alexandra Olteanu and Nicholas J Pangakis and Stefanie Reed and Emily Sheng and Dan Vann and Jennifer Wortman Vaughan and Matthew Vogel and Hannah Washington and Abigail Z. Jacobs},
booktitle={Forty-second International Conference on Machine Learning Position Paper Track},
year={2025},
url={https://openreview.net/forum?id=1ZC4RNjqzU}
}

@misc{pew-atp-wave54-2019,
  author       = {{Pew Research Center}},
  title        = {{American Trends Panel Wave 54}},
  year         = {2020},
  month        = jan,
  day          = {9},
  howpublished = {\url{https://www.pewresearch.org/dataset/american-trends-panel-wave-54/}},
  note         = {Fielded September 16--29, 2019. Topic: Economic inequality.}
}

@misc{qin2026restoringheterogeneityllmbasedsocial,
      title={Restoring Heterogeneity in {LLM}-based Social Simulation: An Audience Segmentation Approach}, 
      author={Xiaoyou Qin and Zhihong Li and Xiaoxiao Cheng},
      year={2026},
      eprint={2604.06663},
      archivePrefix={arXiv},
      primaryClass={cs.CY},
      url={https://arxiv.org/abs/2604.06663}, 
}

@article{shanahan_role_2023,
	title = {Role play with large language models},
	volume = {623},
	issn = {1476-4687},
	url = {https://doi.org/10.1038/s41586-023-06647-8},
	doi = {10.1038/s41586-023-06647-8},
	number = {7987},
	journal = {Nature},
	author = {Shanahan, Murray and McDonell, Kyle and Reynolds, Laria},
	month = nov,
	year = {2023},
	pages = {493--498},
}

@inproceedings{10.1145/3805689.3806444,
author = {D'Alonzo, Samantha and Kreuter, Frauke and Booth, Serena},
title = {Helpful, Harmless, Honest? {RLHF} as Survey Design and Content Moderation},
year = {2026},
isbn = {9798400725968},
publisher = {Association for Computing Machinery},
address = {New York, NY, USA},
url = {https://doi.org/10.1145/3805689.3806444},
doi = {10.1145/3805689.3806444},
booktitle = {Proceedings of the 2026 ACM Conference on Fairness, Accountability, and Transparency},
pages = {7587–7603},
numpages = {17},
location = {
},
series = {FAccT '26}
}

@inproceedings{
poole-dayan2026benchmarking,
title={Benchmarking {Overton} Pluralism in {LLM}s},
author={Elinor Poole-Dayan and Jiayi Wu and Taylor Sorensen and Jiaxin Pei and Michiel A. Bakker},
booktitle={The Fourteenth International Conference on Learning Representations},
year={2026},
url={https://openreview.net/forum?id=f2VxF4QIx1}
}

\appendix

\section{Appendix}
\label{sec:appendix}

\FloatBarrier
\subsection{Emulation Setup}
\label{sec:appendix-emulation-setup}

\subsubsection{LLM Judge}
\label{sec:appendix-emulation-llm-judge}

We validate the LLM judge used in the emulation pipeline against human annotations on a sample of 200 response-category pairs. The human annotator is a member of the research team who had understanding of the research goals, in order to best capture construct validity \citep{wallach2025position}. On the full mapping task, which includes responses judged as unmappable, the judge achieves Cohen's \(\kappa=0.66\) with 74\% raw agreement. Restricting to the 146 pairs where both the judge and human annotator assigned a letter category, agreement rises to \(\kappa=0.72\) (81\% raw agreement). The judge's prompt can be seen in \Cref{fig:judge_prompt}.

\begin{figure}[h]
\centering
\begin{promptbox}
\small
Read the survey response and map it to the best matching multiple-choice option.
You are interpreting a language model's response; after the "answer" it may appear very messy (e.g. interview continuation).
If the first part of a response is at all usable, i.e. looks like an answer continuation, map it, even if the text afterwords frames it in a nonstandard way.
Return only a single option letter.
If the response is unusable, return X.
Interpret the answer at face value; do not try to reason about it or rationalize intentions. Do not infer a stronger or more socially desirable answer than what is actually stated.

\medskip
\textbf{Question:} \emph{[question text]}

\medskip
\textbf{Options:} \emph{[options]}

\medskip
\textbf{Response:} \emph{[response text]}
\end{promptbox}
\caption{Judge prompt used to map free-form model responses to multiple-choice survey options.}
\label{fig:judge_prompt}
\end{figure}

\FloatBarrier
\subsubsection{Base Model Emulation Scaffolding}
\label{appendix-base-model-emulation}

Results are computed across \(N=100\) stochastic generations per question. The unconditioned completion prompt can be seen in \Cref{fig:base_emulation_scaffold_marginal}, and the conditioned completion prompt can be seen in \Cref{fig:base_emulation_scaffold_conditional}.

\begin{table*}[t]
\centering
\small
\caption{Percent usable responses for base open-response models across the seven W54 evaluation settings. Each entry reports the average percentage of generations that the judge successfully mapped to one of the survey answer options, averaged over questions within that setting. Higher values indicate that a larger share of free-form generations could be used in evaluation.}
\label{tab:table_emulation_response_mass}
\begin{tabular}{lrrrrrrrr}
\toprule
Model & Marginal & Democrat & Republican & Very Lib. & Very Cons. & Protestant & Upper Inc. & Mean \\
\midrule
Qwen3-14B Base & 80 & 84 & 82 & 84 & 83 & 82 & 83 & 82 \\
Olmo-3-1025-7B & 79 & 81 & 78 & 82 & 78 & 75 & 81 & 79 \\
Olmo-3-1125-32B & 82 & 87 & 86 & 90 & 86 & 86 & 88 & 87 \\
\bottomrule
\end{tabular}
\end{table*}

\begin{figure}[h!]
\centering
\begin{promptbox}
\small
INTERVIEWER: How would you describe your household's financial situation?\textbackslash{}nPARTICIPANT:

\end{promptbox}
\caption{Prompt scaffold for unconditioned base-model emulation. The model is framed as a participant in an interview-style dialogue and generates a free-form survey response without access to answer options.}
\label{fig:base_emulation_scaffold_marginal}
\end{figure}

\begin{figure}[h!]
\centering
\begin{promptbox}
\small
INTERVIEWER: How would you describe yourself politically?\textbackslash{}nPARTICIPANT: I am a Democrat.\textbackslash{}nINTERVIEWER: How would you describe your household's financial situation?\textbackslash{}nPARTICIPANT:

\end{promptbox}
\caption{Prompt scaffold for demographically conditioned base-model emulation. Demographic conditioning is introduced through a preceding dialogue turn before the survey question is asked.}
\label{fig:base_emulation_scaffold_conditional}
\end{figure}

\FloatBarrier
\subsubsection{Post-trained Model Emulation Scaffolding}

Similar to the base model setup, results are computed across \(N=100\) stochastic generations per question. The post-trained instruction prompts can be seen in \Cref{fig:post-trained_emulation_scaffold_marginal} and \Cref{fig:post-trained_emulation_scaffold_conditional}.

\begin{figure}[h!]
\centering
\begin{promptbox}
\small
\texttt{<system>}You are simulating a participant's answer to a survey question. Generate responses that reflect the full range of American opinion. Return only the participant's answer in plain text, with no markdown or explanation.

\medskip
\texttt{<user>}Survey question: How would you describe your household's financial situation?

\end{promptbox}
\caption{Prompt scaffold for unconditioned post-trained model emulation. The system instruction frames the model as a survey participant and requests a plain-text response to the survey question.}
\label{fig:post-trained_emulation_scaffold_marginal}
\end{figure}

\begin{figure}[h!]
\centering
\begin{promptbox}
\small
\texttt{<system>}You are simulating a participant's answer to a survey question. Generate the response that a person with the following attribute would give. Attribute: <attribute>. Return only the participant's answer in plain text, with no markdown or explanation.

\medskip
\texttt{<user>}Survey question: How would you describe your household's financial situation?

\end{promptbox}
\caption{Prompt scaffold for demographically conditioned post-trained model emulation. The system instruction specifies a target demographic attribute and asks the model to generate a corresponding participant response.}
\label{fig:post-trained_emulation_scaffold_conditional}
\end{figure}

\FloatBarrier
\subsubsection{Self-similarity Comparison}

\Cref{tab:open-response-jaccard} displays the unigram and bigram textual self-similarity across the base and post-trained models.

\begin{table}[h!]
\centering
\small
\caption{Average pairwise Jaccard similarity for open-response generations, averaged across 59 questions and 7 W54 conditions. Responses are normalized and truncated to the first 25 words.}
\label{tab:open-response-jaccard}
\begin{tabular}{lrr}
\toprule
Model & Unigram & Bigram \\
\midrule
Qwen3-14B Base & 0.076 & 0.017 \\
Qwen3-14B & 0.368 & 0.238 \\
\midrule
Olmo-3-1025-7B & 0.089 & 0.014 \\
Olmo-3-7B-Instruct & 0.243 & 0.142 \\
\midrule
Olmo-3-1125-32B & 0.087 & 0.020 \\
Olmo-3.1-32B-Instruct & 0.274 & 0.159 \\
\bottomrule
\end{tabular}
\end{table}

\FloatBarrier
\subsection{Estimation Setup}
\label{sec:appendix-estimation-setup}

\subsubsection{Base Model Estimation Scaffolding}

The completion prompt given to base models for the estimation task can be seen in \Cref{fig:base_estimation_scaffold}.

\begin{figure}[h!]
\centering
\begin{promptbox}
\small
The following is the estimated distribution of <population\_text> to the question: ``<question\_text>''\textbackslash{}nOptions: <options>\textbackslash{}nEstimated distribution: \{\{``A'': 

\end{promptbox}
\caption{Prompt scaffold for base-model estimation. The model directly completes a partially specified probability distribution over survey response options.}
\label{fig:base_estimation_scaffold}
\end{figure}

\FloatBarrier
\subsubsection{Post-trained Model Estimation Scaffolding}

The instruction prompt given to post-trained models for the estimation task can be seen in \Cref{fig:post-trained_estimation_scaffold}.

\begin{figure}[h!]
\centering
\begin{promptbox}
\small
\texttt{<system>}You are estimating the probability distribution of human answers to a multiple-choice survey question. Return exactly one JSON object and nothing else. Use every option letter exactly once as a key. Each value must be an integer percentage. The percentages must sum to 100. Do not include explanation, markdown, code fences, comments, or any keys other than the option letters.

\medskip
\texttt{<user>}Target demographic: estimate the distribution for respondents matching these filters, not for the general population: <target\_demographic>

\medskip
Question: <question\_text>

\medskip
Options: <options>

\medskip
Return a JSON object with integer percentages summing to 100.

\end{promptbox}
\caption{Prompt scaffold for post-trained model estimation. The model is instructed to explicitly estimate the probability distribution of survey responses and return a JSON object with integer percentages summing to 100.}
\label{fig:post-trained_estimation_scaffold}
\end{figure}

\FloatBarrier
\subsection{Metric Definitions}
\label{sec:appendix-metric-definitions}

Let \(Q\) be the set of completed questions included in evaluation, and let \(N = |Q|\).
For each question \(q \in Q\), let:

\begin{itemize}
    \item \(H_q\) be the human response distribution over the options for question \(q\)
    \item \(M_q\) be the model response distribution over the options for question \(q\)
    \item \(H_q(i)\) and \(M_q(i)\) be the mass assigned to option \(i\) by \(H_q\) and \(M_q\) respectively
    \item \(\mathcal{O}_q = \{1, \dots, K_q\}\) be the option set for question \(q\)
\end{itemize}

\subsubsection{Total Variation Distance}

TVD is the largest difference in probability the two distributions assign to any event, and it ranges from 0 (identical) to 1 (disjoint support). It treats all option-level disagreements equally, regardless of how far apart the options are in the answer ordering. For each question \(q\),
\[
\mathrm{TVD}(H_q, M_q)
= \frac{1}{2}\sum_{i \in \mathcal{O}_q} \left|H_q(i) - M_q(i)\right|.
\]

\noindent The pipeline-level score is
\[
\mathrm{MeanTVD}
= \frac{1}{N}\sum_{q \in Q} \mathrm{TVD}(H_q, M_q).
\]

\subsubsection{Wasserstein Distance}

Wasserstein distance is the minimum mass-weighted distance required to transform one distribution into the other, measured in option positions. Placing mass one option away from the correct response is penalized less than placing it three options away. We compute the Wasserstein-1 distance over the ordered answer options per question from cumulative mass differences:

\[
W_1(H_q, M_q)
= \sum_{i=1}^{K_q-1}
\left|
\sum_{j=1}^{i}\bigl(H_q(j) - M_q(j)\bigr)
\right|.
\]

\noindent The pipeline-level score is
\[
\mathrm{MeanWasserstein}
= \frac{1}{N}\sum_{q \in Q} W_1(H_q, M_q).
\]

\subsection{Per-Condition Breakdowns}
\label{sec:appendix-per-condition-breakdowns}

\begin{figure}[h!]
    \centering
    \includegraphics[width=\columnwidth]{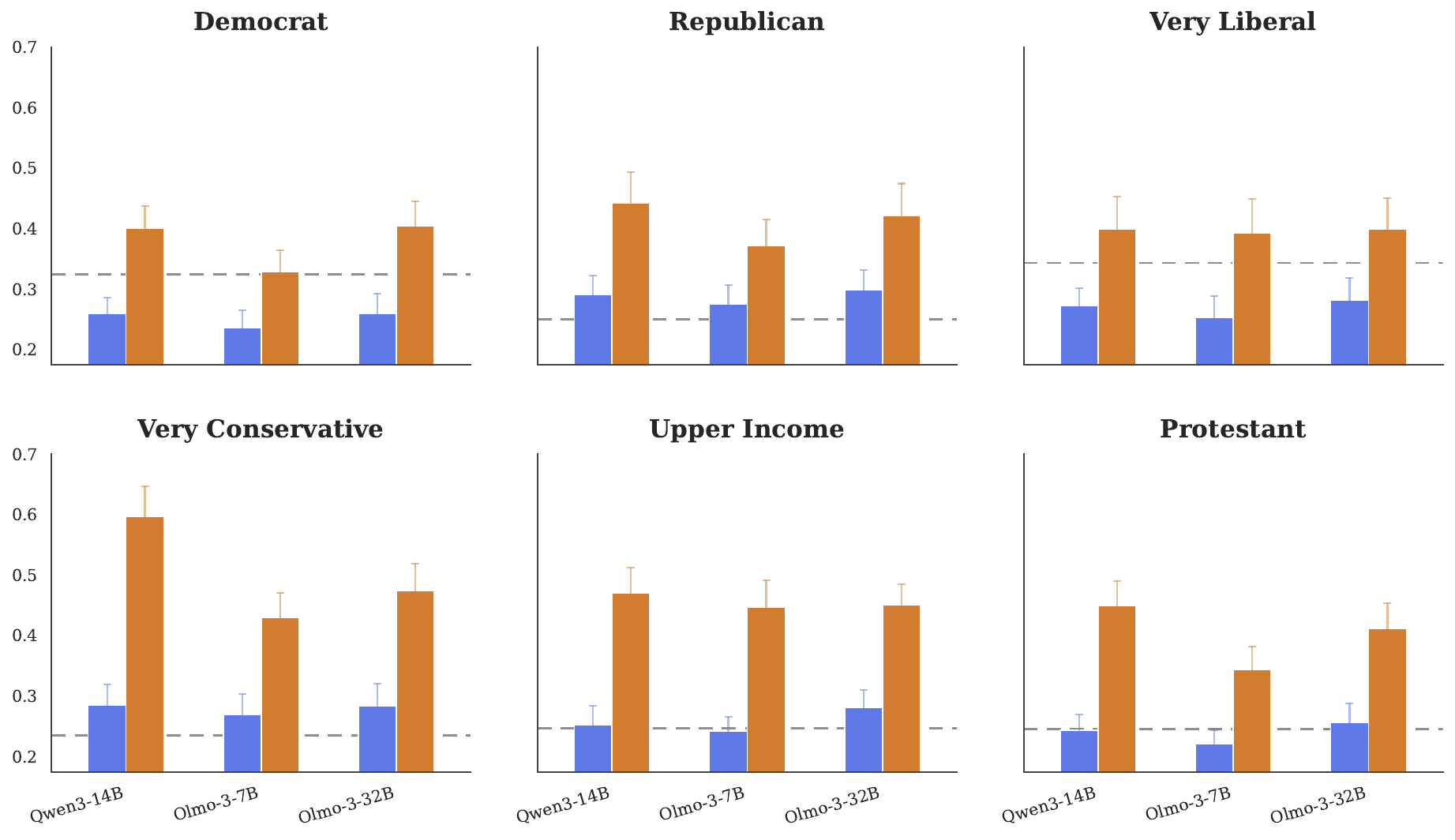}
    \caption{Per-condition emulation error (TVD). Lower is better. Base models have a consistent advantage over post-trained models.}
    \label{fig:conditioned_tvd_results}
\end{figure}

The per-condition emulation error (TVD) can be seen in \Cref{fig:conditioned_tvd_results}. See \Cref{tab:table_emulation_per_condition}  and \Cref{tab:table_estimation_per_condition} for full per-condition emulation and estimation results respectively.

\begin{table*}[t]
\centering
\small
\caption{Emulation per condition}
\label{tab:table_emulation_per_condition}
\resizebox{\textwidth}{!}{%
\begin{tabular}{lrrrrrrrrrrrrrr}
\toprule
Model & \multicolumn{2}{c}{Marginal} & \multicolumn{2}{c}{Democrat} & \multicolumn{2}{c}{Republican} & \multicolumn{2}{c}{Very Lib.} & \multicolumn{2}{c}{Very Cons.} & \multicolumn{2}{c}{Protestant} & \multicolumn{2}{c}{Upper Inc.} \\
\cmidrule(lr){2-3} \cmidrule(lr){4-5} \cmidrule(lr){6-7} \cmidrule(lr){8-9} \cmidrule(lr){10-11} \cmidrule(lr){12-13} \cmidrule(lr){14-15}
 & TVD & Wasserstein & TVD & Wasserstein & TVD & Wasserstein & TVD & Wasserstein & TVD & Wasserstein & TVD & Wasserstein & TVD & Wasserstein \\
\midrule
Qwen3-14B Base & 0.237 & 0.398 & 0.260 & 0.469 & 0.290 & 0.490 & 0.273 & 0.492 & 0.284 & 0.503 & 0.243 & 0.417 & 0.252 & 0.427 \\
Qwen3-14B & 0.447 & 0.648 & 0.400 & 0.564 & 0.442 & 0.688 & 0.400 & 0.611 & 0.597 & 1.021 & 0.450 & 0.658 & 0.470 & 0.734 \\
\midrule
Olmo-3-1025-7B & 0.236 & 0.386 & 0.237 & 0.402 & 0.276 & 0.453 & 0.253 & 0.443 & 0.270 & 0.454 & 0.222 & 0.372 & 0.242 & 0.397 \\
Olmo-3-7B-Instruct & 0.290 & 0.426 & 0.329 & 0.492 & 0.371 & 0.623 & 0.393 & 0.636 & 0.430 & 0.656 & 0.344 & 0.496 & 0.447 & 0.683 \\
\midrule
Olmo-3-1125-32B & 0.253 & 0.389 & 0.260 & 0.427 & 0.299 & 0.483 & 0.281 & 0.488 & 0.284 & 0.463 & 0.257 & 0.408 & 0.281 & 0.441 \\
Olmo-3.1-32B-Instruct & 0.429 & 0.647 & 0.404 & 0.592 & 0.421 & 0.640 & 0.399 & 0.611 & 0.475 & 0.733 & 0.412 & 0.611 & 0.450 & 0.667 \\
\midrule
uniform reference distribution & 0.251 & 0.547 & 0.325 & 0.712 & 0.250 & 0.503 & 0.343 & 0.768 & 0.235 & 0.498 & 0.245 & 0.525 & 0.247 & 0.520 \\
\bottomrule
\end{tabular}%
}
\end{table*}

\begin{table*}[t]
\centering
\small
\caption{Estimation per condition}
\label{tab:table_estimation_per_condition}
\resizebox{\textwidth}{!}{%
\begin{tabular}{lrrrrrrrrrrrrrr}
\toprule
Model & \multicolumn{2}{c}{Marginal} & \multicolumn{2}{c}{Democrat} & \multicolumn{2}{c}{Republican} & \multicolumn{2}{c}{Very Lib.} & \multicolumn{2}{c}{Very Cons.} & \multicolumn{2}{c}{Protestant} & \multicolumn{2}{c}{Upper Inc.} \\
\cmidrule(lr){2-3} \cmidrule(lr){4-5} \cmidrule(lr){6-7} \cmidrule(lr){8-9} \cmidrule(lr){10-11} \cmidrule(lr){12-13} \cmidrule(lr){14-15}
 & TVD & Wasserstein & TVD & Wasserstein & TVD & Wasserstein & TVD & Wasserstein & TVD & Wasserstein & TVD & Wasserstein & TVD & Wasserstein \\
\midrule
Qwen3-14B Base & 0.194 & 0.419 & 0.282 & 0.591 & 0.214 & 0.426 & 0.300 & 0.626 & 0.237 & 0.493 & 0.223 & 0.487 & 0.203 & 0.403 \\
Qwen3-14B & 0.193 & 0.407 & 0.208 & 0.418 & 0.227 & 0.446 & 0.223 & 0.430 & 0.258 & 0.482 & 0.198 & 0.402 & 0.187 & 0.370 \\
\midrule
Olmo-3-1025-7B & 0.235 & 0.529 & 0.327 & 0.720 & 0.250 & 0.520 & 0.365 & 0.795 & 0.268 & 0.582 & 0.241 & 0.535 & 0.258 & 0.556 \\
Olmo-3-7B-Instruct & 0.175 & 0.351 & 0.318 & 0.698 & 0.256 & 0.523 & 0.387 & 0.866 & 0.370 & 0.760 & 0.255 & 0.538 & 0.236 & 0.483 \\
\midrule
Olmo-3-1125-32B & 0.206 & 0.432 & 0.260 & 0.555 & 0.197 & 0.379 & 0.273 & 0.578 & 0.210 & 0.419 & 0.200 & 0.422 & 0.212 & 0.432 \\
Olmo-3.1-32B-Instruct & 0.170 & 0.335 & 0.191 & 0.376 & 0.181 & 0.333 & 0.216 & 0.401 & 0.215 & 0.370 & 0.153 & 0.293 & 0.169 & 0.321 \\
\midrule
Claude Opus 4.6 & 0.130 & 0.245 & 0.124 & 0.224 & 0.164 & 0.311 & 0.119 & 0.195 & 0.174 & 0.342 & 0.141 & 0.286 & 0.143 & 0.266 \\
\midrule
uniform baseline & 0.251 & 0.547 & 0.325 & 0.712 & 0.250 & 0.503 & 0.343 & 0.768 & 0.235 & 0.498 & 0.245 & 0.525 & 0.247 & 0.520 \\
\bottomrule
\end{tabular}%
}
\end{table*}

\FloatBarrier
\subsection{First-token Extraction}
\label{sec:appendix-first-token-extraction}

\begin{figure}[h!]
  \centering
\includegraphics[width=0.9\linewidth]{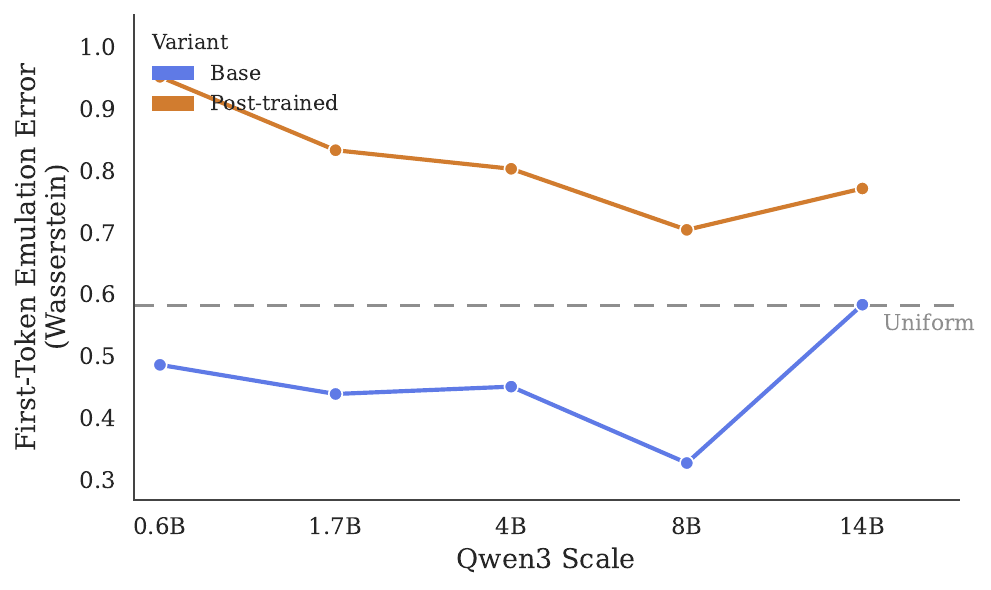} 
  \caption {First-token emulation error (Wasserstein distance, averaged across 7 evaluation conditions) for Qwen3 base and post-trained models from 0.6B to 14B parameters. Base models outperform their post-trained counterparts at every scale, with the strongest performance at 8B. The 14B base model degrades to near-uniform levels, motivating our adoption of open-response emulation for the main evaluation.}
  \label{fig:qwen_first-token-wasserstein}
\end{figure}

Scaffolds were selected to maximize probability mass over valid answer tokens, independent of distributional fit to human data. The base model prompt ends with ``My choice is Letter:'' to constrain generation to answer tokens. The  post-trained model uses a system prompt framing the model as a survey participant.

First-token extraction shows a consistent base advantage across scales (\Cref{fig:qwen_first-token-wasserstein}), with the strongest performance at 8B parameters. Performance degrades at 14B, potentially due to stronger positional priors. Models at this scale show evidence of positional-bias confounding, motivating our adoption of open-response emulation for the main evaluation. The directional finding that base models outperform instruction-tuned models is consistent with the open-response emulation results reported in the main paper.

\end{document}